\documentclass[letterpaper, 10 pt, conference]{ieeeconf}  

\IEEEoverridecommandlockouts                              

\usepackage{graphicx}
\usepackage{tabularx}
\graphicspath{ {./images/} }
\usepackage{amsmath}
\DeclareMathOperator{\sgn}{sgn}
\usepackage{float} 
\usepackage{hyperref}
\usepackage{nameref}

\title{\LARGE \bf
Series Elastic Force Control for Soft Robotic Fluid Actuators
}

\author{Chunpeng Wang$^{1}$ and John P. Whitney$^{1,2}$%
\thanks{$^{1}$The authors are with the department of Mechanical and Industrial Engineering, Northeastern University, Boston, MA 02115, USA.}%
\thanks{$^{2}$corresponding author,
        {\tt\small j.whitney@northeastern.edu}}%
}

\begin{document}

\maketitle
\thispagestyle{empty}
\pagestyle{empty}

\begin{abstract}

Fluid-based soft actuators are an attractive option for lightweight and human-safe robots.  These actuators, combined with fluid pressure force feedback, are in principle a form of series-elastic actuation (SEA), in which nearly all driving-point (e.g. motor/gearbox) friction can be eliminated. Fiber-elastomer soft actuators offer unique low-friction and low-hysteresis mechanical properties which are particularly suited to force-control based on internal pressure force feedback, rather than traditional external force feedback using force/tactile sensing, since discontinuous (Coulomb) endpoint friction is unobservable to internal fluid pressure. However, compensation of endpoint smooth hysteresis through a model-based feedforward term is possible.  We report on internal-pressure force feedback through a disturbance observer (DOB) and model-based feedforward compensation of endpoint friction and nonlinear hysteresis for a 2-DOF lightweight robotic gripper driven by rolling-diaphragm linear actuators coupled to direct-drive brushless motors, achieving an active low-frequency endpoint impedance range (``Z-width") of 50dB.

\end{abstract}

\section{INTRODUCTION}
In many situations, being able to vary the endpoint stiffness of a robotic manipulator over a wide range is highly desirable---stiff when high bandwidth performance is needed, and soft when interacting with a delicate environment or when visual feedback is poor or unavailable~\cite{ham2009compliant}.  Haptic interfaces, in particular, seek to maximize the ratio between maximum and minimum endpoint impedance (``Z-width'')~\cite{colgate_issues_1995}, giving the ability to render the widest possible range of virtual environments\footnote{In this paper, we use the term ``impedance'' in both the formal sense (mechanical impedance for linear systems is defined as $Z(s)=F(s)/V(s)$, where $s$ is the Laplace variable and $F$ and $V$ are the port force and velocity respectively) and the informal sense, referring simultaneously and variably to stiffness, friction, mass, and damping.  In informal usage, ``low impedance'' usually refers to minimizing the stiffness and friction which dominate low-frequency characteristics of a mechanical system.}.  A large Z-width is also desirable for rehabilitation robots, which would ideally offer a continuously variable impedance from peak virtual stiffness down to a state of perfect backdrivability (zero impedance); low-impedance operations are generally useful for safety-critical and co-robotics applications~\cite{zinn2004playing,park2009safe}.

Two classic, complementary approaches to this challenge are the use of high-performance mechanical transmissions to offload the actuator mass from the moving mass of the manipulator to the base, as exemplified by the Whole Arm Manipulator~\cite{townsend1999teleoperator}, and high-bandwidth endpoint force-feedback (e.g. admittance control) which is designed to render a specific desired endpoint impedance and actively compensate for sources of mechanical friction.  A long-recognized limitation is that force-feedback cannot significantly reduce the physical endpoint inertia without jeopardizing the closed-loop system stability~\cite{whitney_historical_1985,eppinger1987introduction,chae_an_dynamic_1987}.  However, this restriction can be circumvented if purposeful physical compliance (so-called ``series elasticity'') is added between the actuator and the endpoint~\cite{pratt.95,hurst2010actuator}.

Fluid-powered actuators, and soft fluid actuators in particular~\cite{majidi2014soft}, have inherently high force density~\cite{torquedensity_1991}, allowing for low endpoint mass even for serial-chain manipulators with many degrees of freedom.  By adjusting the volumetric compliance and internal geometry of the transmission hoses, a purposeful series elasticity and damping can be tuned~\cite{sivaselvan2008dynamic}.  The measurement of the internal pressure in the hydraulic or pneumatic line affords an estimate of the joint torques.  However, any static friction at the endpoint (e.g. hydraulic seal friction) will not be observable from internal pressure measurements.   Soft-continuum and diaphragm-type fluid actuators based on material elasticity and/or non-rubbing diaphragm seals exhibit nearly-zero static friction, allowing external interaction forces to be estimated with high precision using only internal fluid pressure measurements and without any requirement for electrical wiring to the endpoint.

In this paper we investigate fluid-actuated robotic systems with purposeful fluid/hose compliance~\cite{whitney2016hybrid}, internal pressure force-feedback, and low-friction fiber-elastomer soft actuators.  Section II describes the exemplary system, a 2-DOF lightweight gripper~\cite{schwarm_floating-piston_2019} with details on fluid pressure force feedback setup, and system identification results.  Section III describes the disturbance observer (DOB) force feedback method, adapted to the case of fluid-actuated systems, and Section IV describes a model-based approach to compensating for smooth endpoint friction without any endpoint state or external force sensing, and presents detailed Z-width measurements.  The accompanying video shows operation of the gripper, passive behavior, and the qualitative minimum and maximum impedance performance.

\section{SYSTEM DESIGN AND MODELLING}

\begin{figure}[thpb]
    \centering
    \framebox{\parbox{3in}{\includegraphics[width=3in]{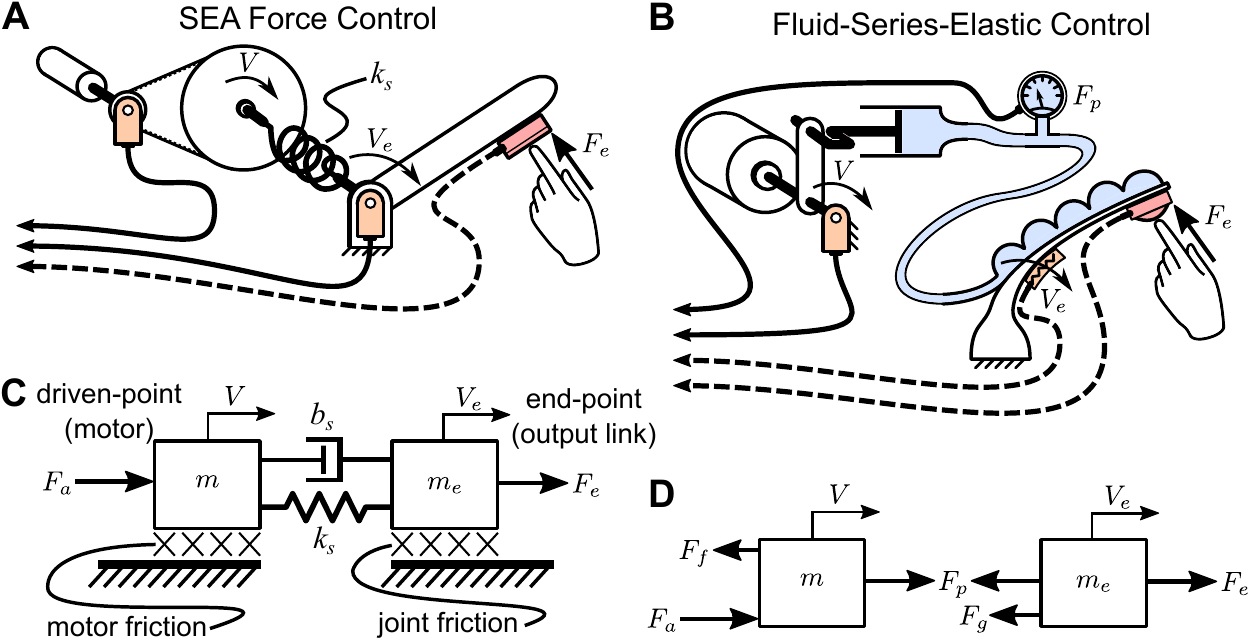}}}
    \caption{Force feedback systems, with primary feedback signals (solid) and secondary feedback signals (dashed). 
    {\bf(A)} Series-elastic configuration, where purposeful compliance allows an estimate of internal torque using dual encoder feedback. {\bf(B)} Series-elastic system with series fluid connection between motor and endpoint link. {\bf(C--D)} 2-DOF series-elastic system model with force and parameter definitions. All the parameters and variables here are transferred into motor rotation frame with SI unit. The translation figure of model is used to represent rotational model for simplicity.}  
    \label{fig:configs}
\end{figure}

Maximizing the dynamic range of a manipulator's endpoint impedance requires a collaboration between the physical hardware and closed-loop feedback control design. Fig.~\ref{fig:configs}A/B shows two different approaches, where the primary sensory feedback signals are shown with solid lines, and less-commonly employed feedback signals are shown dashed.  $V_e$ and $V$ are the external (endpoint) and driving-point (motor) velocities, $F_p$ is the transmission internal force, and $F_e$ is the external (endpoint) force.  Fig.~\ref{fig:configs}C/D shows models of traditional SEA system and fluid-series-elastic system, where $F_a$ is the applied motor force, $F_f$ is the motor friction, $F_g$ is the endpoint joint friction, $F_p$ is hydraulic pressure force, $m$ and $m_e$ are the motor and endpoint inertia, and $k_s$ and $b_s$ are the series stiffness and damping.  In this linear SEA model, driving-point friction/damping is modeled as $F_f=bV + kV/s$ and endpoint friction is $F_g=b_eV_e+k_eV_e/s$, where $b$ and $k$ are motor damping and stiffness, and $b_e$ and $k_e$ are joint damping and stiffness\footnote{Note: In this paper, our equations and models use the translational lumped-element equivalent system convention, while the experimental system employed is a rotational system.  References to forces and velocities are given/plotted in the corresponding rotational (torque/angular velocity) units, and presented in the motor frame unless otherwise noted.  Internal fluid pressure is converted to the equivalent internal force, $F_p$, which for the experimental system is reported as an internal torque acting on the motor, in units of Newton-meters.}.

\subsection{Actuation and Control Architecture}
In the impedance control approach~\cite{Hogan-I.85,vanderborght2013variable}, a low-friction actuator drives a joint via a stiff series connection; driving-point state feedback is used to render the desired endpoint impedance, and the stiff transmission ensures that the driving-point impedance is reflected at the endpoint accurately.  However, without force feedback, the system cannot render an endpoint impedance below the passive impedance (i.e. friction and inertia) of the motor/gearbox/actuator.  In the \emph{series elastic actuation (SEA)} approach, as shown in Fig.~\ref{fig:configs}-A, the transmission stiffness is purposefully reduced, easing the requirements on closed-loop bandwidth; as the driving-point and endpoint become increasingly decoupled, it is easier to stably compensate for actuator-side friction and mass~\cite{pratt.95,sensinger2006improvements,paine2014design}.  Achieving maximum performance may entail external force sensing in addition to full-state feedback, but more commonly, the external force is estimated via the internal force $F_p$ in the transmission elastic element as $F_p=\int k(V_e-V)dt$.  Mechanical-spring SEA designs require the actuator and external joint to be located in close proximity, resulting in high endpoint mass for proximal degrees of freedom in serial-chain manipulators.  Fluid-actuated systems (Fig.~\ref{fig:configs}-B) offer a variation on the traditional SEA configuration; with a remote actuator and the mechanical spring replaced by a fluid-filled hose and internal force now measured with a pressure sensor.  The series compliance and damping are tuned by controlling the volumetric stiffness and internal geometry of the hydraulic line~\cite{sivaselvan2008dynamic}, or, in the case of a soft-continuum actuator, modification of the material and geometry of the endpoint actuator.  Soft-material strain and force sensors can be added to these systems at the endpoint, giving high-quality endpoint state and force information~\cite{majidi2014soft}. This eliminates the need to model the internal stiffness and damping properties of the soft actuator, but presents challenges for high-bandwidth control due to sensor-actuator non-collocation---achieving an equivalent level of performance without any endpoint sensing is highly desirable.

\subsection{Mechanical Design and Testbed Setup}

\begin{figure}[thpb]
    \centering
    \framebox{\parbox{3in}{\includegraphics[width=3in]{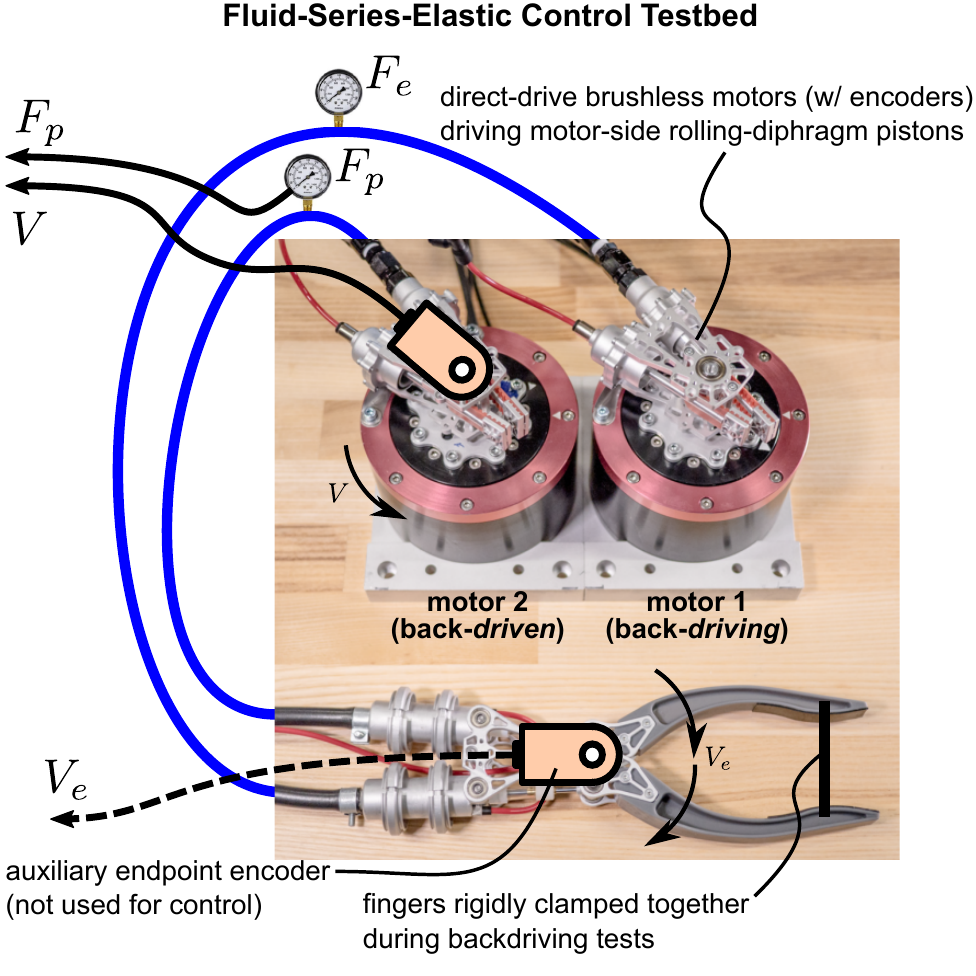}}}
    \caption{Diaphragm-actuated 2-DOF gripper. 
    Testing endpoint impedance by using one finger to backdrive the other. Pressure-to-torque scaling is based on measured cylinder pressure area and actuator geometry.}
    \label{fig:Details of hand-motor-hand system}
\end{figure}

As a testbed for internal fluid pressure force feedback for fluid-actuated systems, we use a rolling-diaphragm-actuated 2-DOF gripper~\cite{schwarm_floating-piston_2019}, shown in Fig.~\ref{fig:Details of hand-motor-hand system}.

Each actuator is a double-acting piston sealed by low friction fiber-elastomer rolling-diaphragms. 
The two diaphragm actuators drive independent fingers, affording wrist pitch and gripper pinch DOFs; the input ends of the transmission are connected to individual direct-drive brushless motors (Akribis ACD120-80) via fiber-reinforced rubber hoses, and the motor is coupled via rotary versions of the diaphragm actuators, introduced in~\cite{whitney2016hybrid}.  This approach combines the low mass and friction of soft fluid actuators with the controllability and performance of direct-drive SEA.

Fluid pressure is recorded using an analog input EtherCAT terminal with 16bit resolution.  A temporary optical encoder made by US DIGITAL with maximum resolution 10,000 pulses/rev (40,000 counts/rev with quadrature) is used to measure the position of the backdriven finger during system identification (not used for control).  
The low-level control code uses the Simple Open EtherCAT Master (SOEM) library with the main control loop running at 2kHz on an isolated CPU core.  More details of testbed setup can be found in~\cite{schwarm_floating-piston_2019}.

\subsection{System Modeling and Identification}

The lumped element model of our system is shown in Fig.~\ref{fig:configs}C/D.
The Laplace-transformed equations of motion can be written as $\left(ms^2+bs+k\right)X=F_p$ for the motor actuator plant, $\left(m_es^2+b_es+k_e\right)X_e=F_e-F_p$ for the endpoint finger plant, and $F_p=\left(b_s s+k_s\right) \left(X_e-X\right)$ for the hydraulic line, where $X$ and $X_e$ are the position of actuator motor and endpoint finger position. 


In Fig.~\ref{fig:Details of hand-motor-hand system} the test conditions are shown for system identification and closed-loop impedance characterization purposes: the two gripper fingers are bonded together and one backdrives the other, serving as an external force/load. In the following, the parameters and values of the endpoint plant are for both fingers.  Endpoint position $X_e$ is only used for system identification and endpoint impedance measurement, but not for closed-loop feedback. 
Note that in this paper we use and report the impedance $Z(s)$ and admittance $Y(s)=Z(s)^{-1}$ under the assumption of a linear system. 

\begin{figure}[thpb]
    \centering
    \framebox{\parbox{3in}{\includegraphics[width=3in]{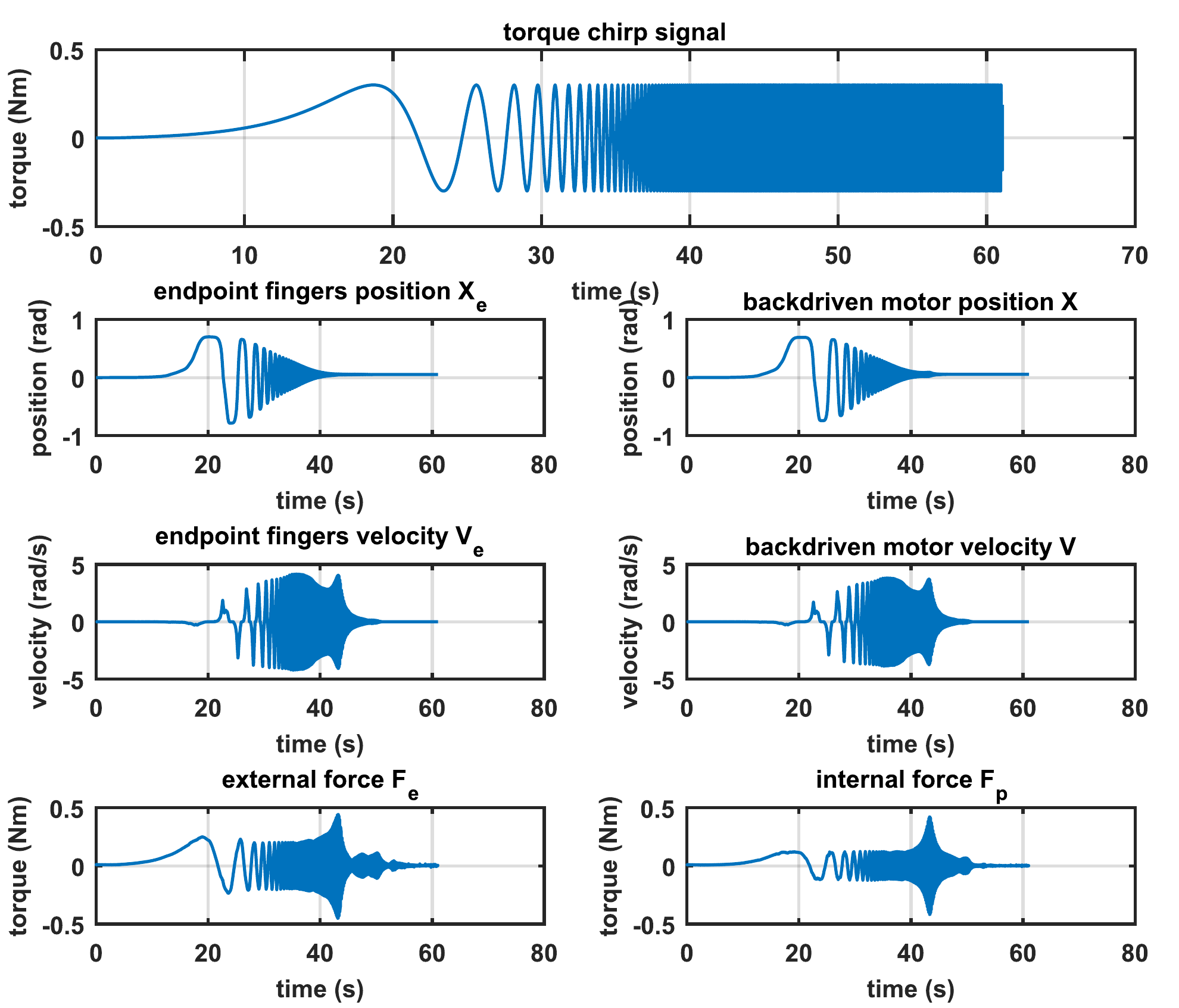}}}
    \caption{Time domain plot of torque chirp signal $T_{chirp}$, endpoint position $X_e$, backdriven motor position $X$, endpoint velocity $V_e$, backdriven motor velocity $V$, external force $F_e$, and internal pressure force $F_p$.}
    \label{fig:chirp}
\end{figure}

Applying a torque chirp signal (where the amplitude is $0.3$Nm and frequency range from $0.01$Hz to $1000$Hz) on driving motor, we collect the data of actuator position $X$ and velocity $V$, endpoint position $X_e$ and velocity $V_e$, internal force $F_p$, and external force $F_e$ shown as Fig.~\ref{fig:chirp}.

\begin{figure}[thpb]
    \centering
    \framebox{\parbox{3in}{\includegraphics[width=3in]{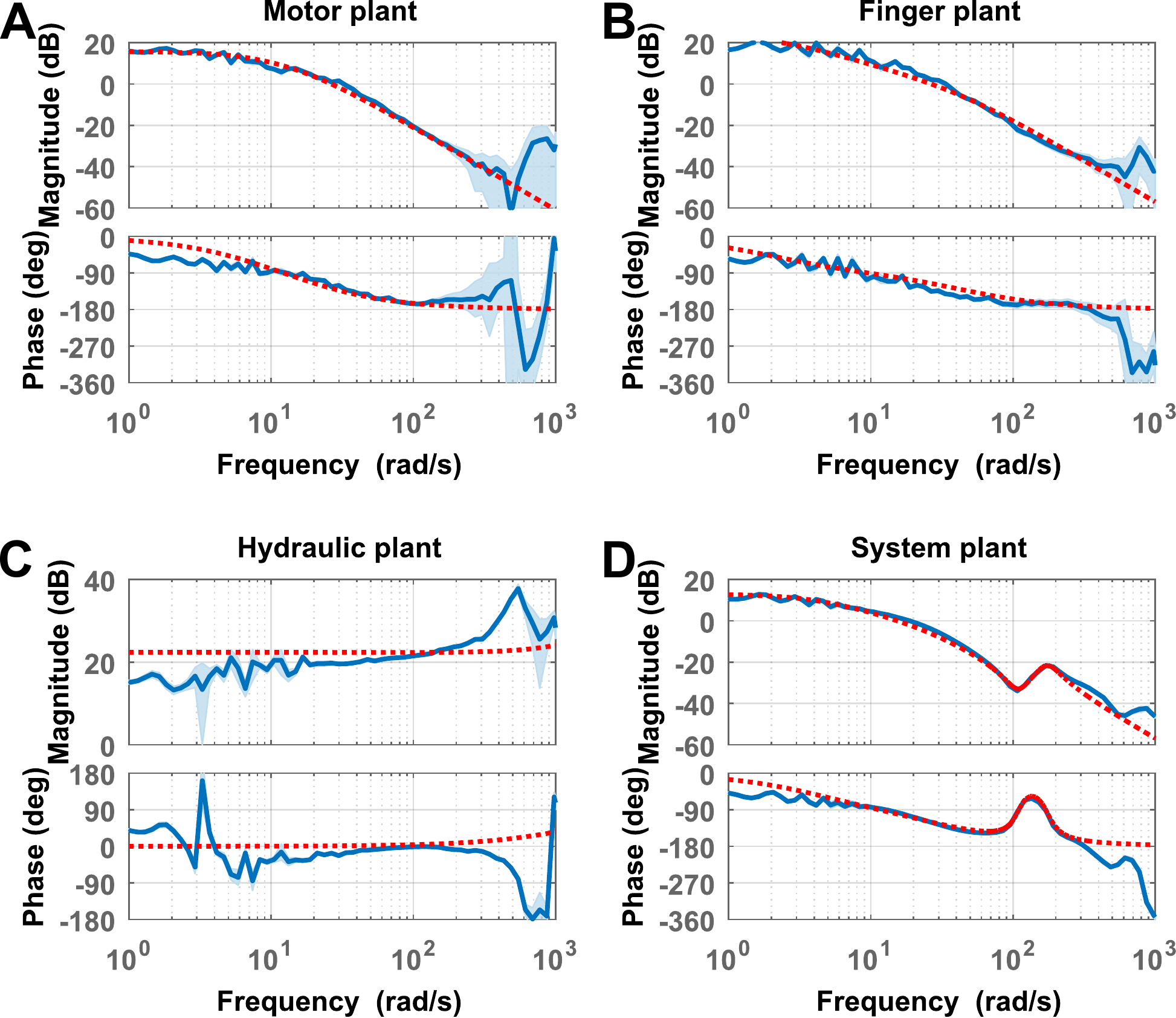}}}
    \caption{System Identification with original passive system (blue) and linear approximated system (red). Fingers in hand are bonded together and one backdrives the other using a logarithmic chirp signal.  The identified frequency response is estimated using the Blackman-Tukey method.  Shaded regions represent \mbox{$1$-$\sigma$} error bands.  {\bf(A)} Motor plant, $X/F_p$. {\bf(B)} Fingers plant, $X_e/(F_e-F_p)$. {\bf(C)} Series-dynamics/hydraulic connection plant, $F_p/(X_e-X)$. {\bf(D)} Endpoint system plant, $X_e/F_e$, where $X_e=V_e/s$ and $X=V/s$.}
    \label{fig:system ID}
\end{figure}

The system identification process is as follows: A set of time-domain data is combined using endpoint finger position $X_e$ as output and external pressure force $F_e$ as input by the \texttt{iddata()} function in MATLAB.  The corresponding set of frequency-domain data is calculated using the time-domain data created above with start frequency of $0.01Hz$ and end frequency of $1000Hz$ by the \texttt{spafdr()} function. The whole system at endpoint is identified using \texttt{tfest()} function with 2 zeros and 4 poles focusing at the range from $0$ to $400rad/s$. The whole system at endpoint is shown in Fig.~\ref{fig:system ID}-D. The parameters of each separate system (Tab.~\ref{tab:identified system parameters}) can be decoupled from the transfer function of the whole system. By comparing the linear approximation of each system, we find all match the original data very well except for the hydraulic line approximation, where a linear damper poorly approximates nonlinear viscous losses.

\begin{table}[h]
    \caption{Identified parameters of system plant.}
    \label{tab:identified system parameters}
    \begin{center}
    \begin{tabular}{|c||c|}
        \hline
        actuator motor inertia $(Nm/(rad/s^2))$ & $1.1116\times 10^{-3}$ \\ 
        \hline
        actuator motor damping $(Nm/(rad/s))$ & $2.9814\times 10^{-2}$ \\ 
        \hline
        actuator motor stiffness $(Nm/rad)$ & $0.1642$ \\
        \hline
        endpoint finger inertia $(Nm/(rad/s^2))$ & $0.7089\times 10^{-3}$ \\
        \hline
        endpoint finger damping $(Nm/(rad/s))$ & $3.3879\times 10^{-2}$ \\
        \hline
        endpoint finger stiffness $(Nm/rad)$ & $0.0637$ \\
        \hline
        hydraulic line damping $(Nm/(rad/s))$ & $9.2453\times 10^{-3}$ \\
        \hline
        hydraulic line stiffness $(Nm/rad)$ & $13.0782$ \\
        \hline
    \end{tabular}
    \end{center}
\end{table}

\section{FORCE FEEDBACK CONTROL}
Traditional model-based impedance control on pneumatic soft actuators~\cite{xiangrong_shen_independent_2005,slightam_sliding_2019} control fluid flow and pressure to render a target impedance, using a velocity source actuator (e.g. fluid servo-valve), in contrast to the present approach, where the electric motor provides a torque source.  
Our electric-hydraulic actuator is designed to exhibit very low mechanical impedance, but as with any system, there is always friction in the joints, transmission, and especially in the motor and motor shaft bearings.  Force feedback control is a simple and direct way to compensate for mechanical impedance~\cite{kurfess_impedance_2004}. Applying simple proportional force feedback, $F_a(t)=K_fF_e(t)$ where $K_f$ is the force feedback constant gain, for a 1-DOF inertia-spring-damper system with nonlinear friction $F_f(x,\dot{x})$, the equation of motion for the closed-loop system is
\begin{equation}
    \label{eqn:simple force feedback}
    m \ddot{x}+b \dot{x} +kx+F_f(x,\dot{x})=F_a+F_e=\left(K_f+1\right)F_e,
\end{equation}
which may be arranged into the equivalent system
\begin{equation}
    \label{eqn:equivalent force feedback}
    \frac{m}{K_f+1}\ddot{x}+\frac{b}{K_f+1}\dot{x}+\frac{k}{K_f+1}x+\frac{F_f(x,\dot{x})}{K_f+1}=F_e.
\end{equation}
The inertia, damping, stiffness, and nonlinear friction of the system are all proportionally reduced as $K_f$ increases. Theoretically, this system can be shown to be passive for any $K_f\geq -1$, where ``passive'' refers to a system that is stable when in contact with any potential environmental impedance~\cite{colgate_analysis_1989}.  However,  every real system has some internal degrees of freedom, whether purposeful (series elastic actuation) or incidental (internal vibration modes), creating a situation of non-collocation between sensing and actuation ports.

\subsection{Internal Force Feedback versus External Force Feedback}
A general 2-DOF series-elastic system is shown in Fig.~\ref{fig:configs}-C. Using linear stiffness and linear damping approximations for nonlinear friction of actuator ($k,b$) and end-effector ($k_e,b_e$), and applying proportional force feedback using the \emph{internal} force $F_a(t)=K_f F_p(t)$, we find the same passivity limit as the 1-DOF case, $K_f\geq -1$, as the endpoint dynamics can be considered part of the passive environment.  However, any DOFs internal to the feedback port for the 1-DOF case or the 2-DOF internal force feedback case will result in the stricter passivity constraint $-1\leq K_f\leq 1$, due to the non-collocation introduced by internal actuator dynamics or internal structural resonance~\cite{colgate_analysis_1989}.  In practice, if the environmental conditions present a known range of impedances, passivity constraints may be strategically violated to achieve lower endpoint impedance.  Care must be taken to prevent destabilizing environmental interactions (e.g. rigid surface contact or coupling to a large inertia).

\begin{figure}[thpb]
    \centering
    \framebox{\parbox{3in}{\includegraphics[width=3in]{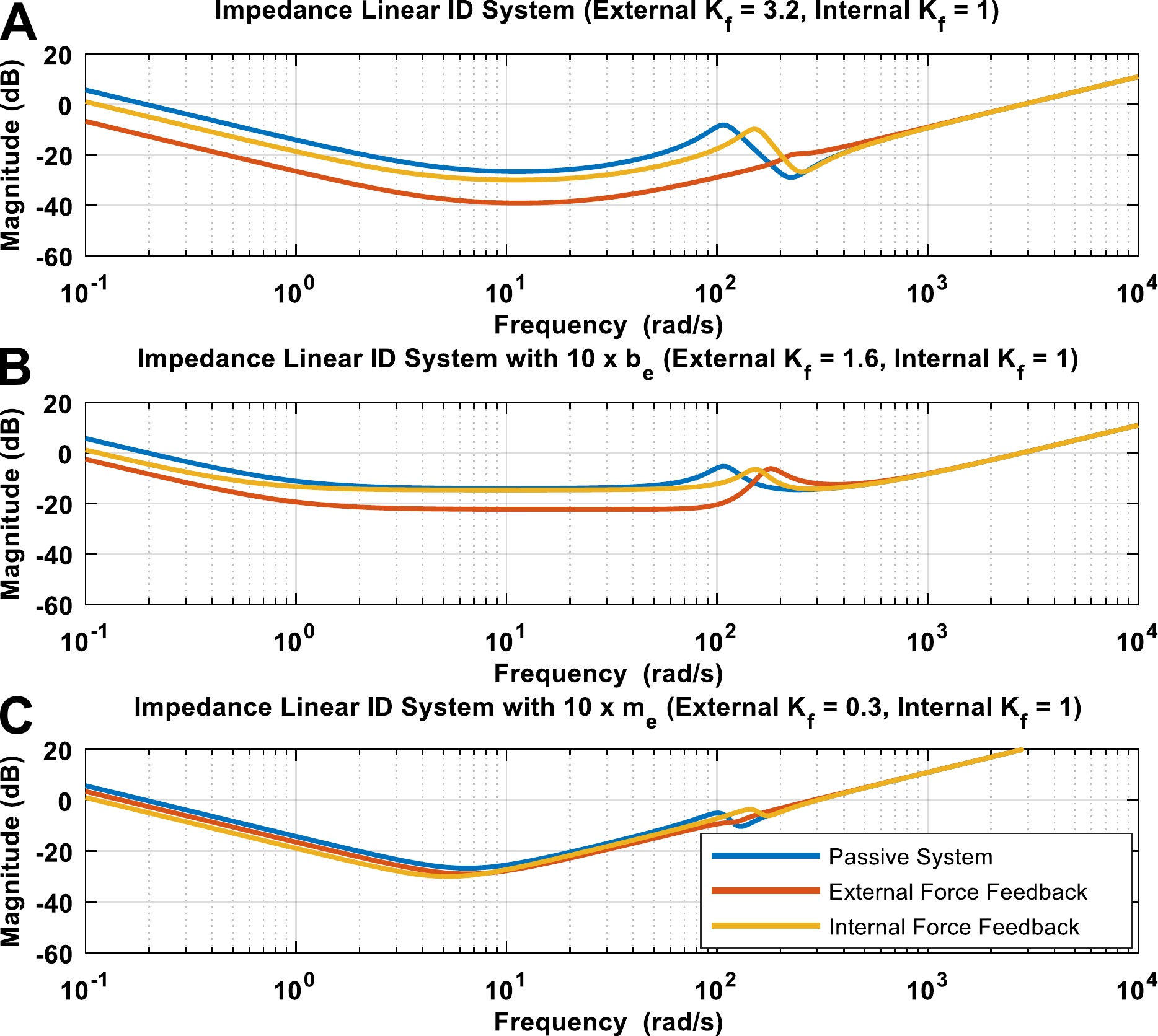}}}
    \caption{Comparison between passive system, internal force feedback, and external force feedback. Endpoint port impedance (lower is better) for the identified model of the passive gripper system (blue), with external force feedback (red), and with internal force feedback when $K_f=1$ (yellow), including endpoint masses, parallel damping, parallel stiffnesses, and transmission compliance and damping. {\bf(A)} Comparison using identified parameters. {\bf(B)} Comparison using identified parameters, except $b_e$ is increased by a factor of ten. {\bf(C)} Comparison using identified parameters, except $m_e$ is increased by a factor of ten. \emph{Maximum passive $K_f$ is employed for all external force feedback cases.}}
    \label{fig:internal and external force feedback admittance comparison}
\end{figure}

Fig.~\ref{fig:internal and external force feedback admittance comparison}-A shows the endpoint impedance passively rendered by both internal and external force feedback, using the identified linear model parameters for the gripper system.  These results are characteristic, showing that external force feedback usually outperforms internal force feedback, at low--mid frequencies.  Internal force feedback is particularly ineffective if the endpoint friction is large (Fig.~\ref{fig:internal and external force feedback admittance comparison}-B), although if endpoint mass is large enough ($m_e >> m$), then internal force feedback is superior (Fig.~\ref{fig:internal and external force feedback admittance comparison}-C).  Note, however, that passivity limits mean that all forms of force feedback are unable to improve high-frequency (inertial) impedance beyond the passive case.

For internal force feedback, using the conservative passivity limit $K_f=1$, we can examine the endpoint impedance in the low frequency limit:
\begin{equation}
    \lim_{s\rightarrow 0}Z(s)\bigg\rvert_{K_f=1} = \frac{(k+2k_e)k_s + kk_e}{(2k_s+k)s}.
\end{equation}
First, consider the case where the actuator is not backdrivable, i.e. $k>>k_s,k_e$.  In this situation, internal force feedback cannot passively reduce the endpoint impedance at all:
\begin{equation}
    \lim_{s\rightarrow 0}Z(s)\bigg\rvert_{K_f=1,\; k>>\{k_s,k_e\}} = \frac{(k_s+k_e)}{s},
\end{equation}
However, if the actuator is backdrivable, such that the transmission stiffness exceeds the driving and endpoint friction, $k_s>>k,k_e$, we find
\begin{equation}
    \lim_{s\rightarrow 0} Z(s)\bigg\rvert_{K_f=1, \; ks>>\{k,k_e\}} = \frac{(\frac{1}{2}k + k_e)}{s},
\end{equation}
indicating that internal force feedback can passively reduce the driving point friction ($k$) by half.

Many methods to improve impedance compensation performance beyond simple proportional force-feedback have been proposed, including \emph{force feedback loop shaping}, where a general force feedback filter replaces the proportional force gain, $K_f\rightarrow K_f(s)$~\cite{buerger2007complementary}, \emph{natural admittance control (NAC)}~\cite{newman_stable_1994}, \emph{model-based force control}~\cite{hart2014absolutely}, and the \emph{disturbance observer (DOB)} control method~\cite{DOB_original}.  These equivalent methods, rather than applying force feedback control effort uniformly across all frequencies, allow high-gain force feedback to target only the lower frequencies where friction, stiffness, and damping of the system dominate, without attempting to reduce inertia at higher frequencies, beyond causality and passivity limits.

\subsection{Disturbance Observer Framework}
Disturbance observers use state measurements and a nominal inverse model of the system plant to estimate the expected external force; this is compared to the actual external force, with the difference becoming an estimate of the disturbance signal, which is then subtracted from the motor input command to remove the disturbance.

\begin{figure}[thpb]
    \centering
    \framebox{\parbox{3in}{\includegraphics[width=3in]{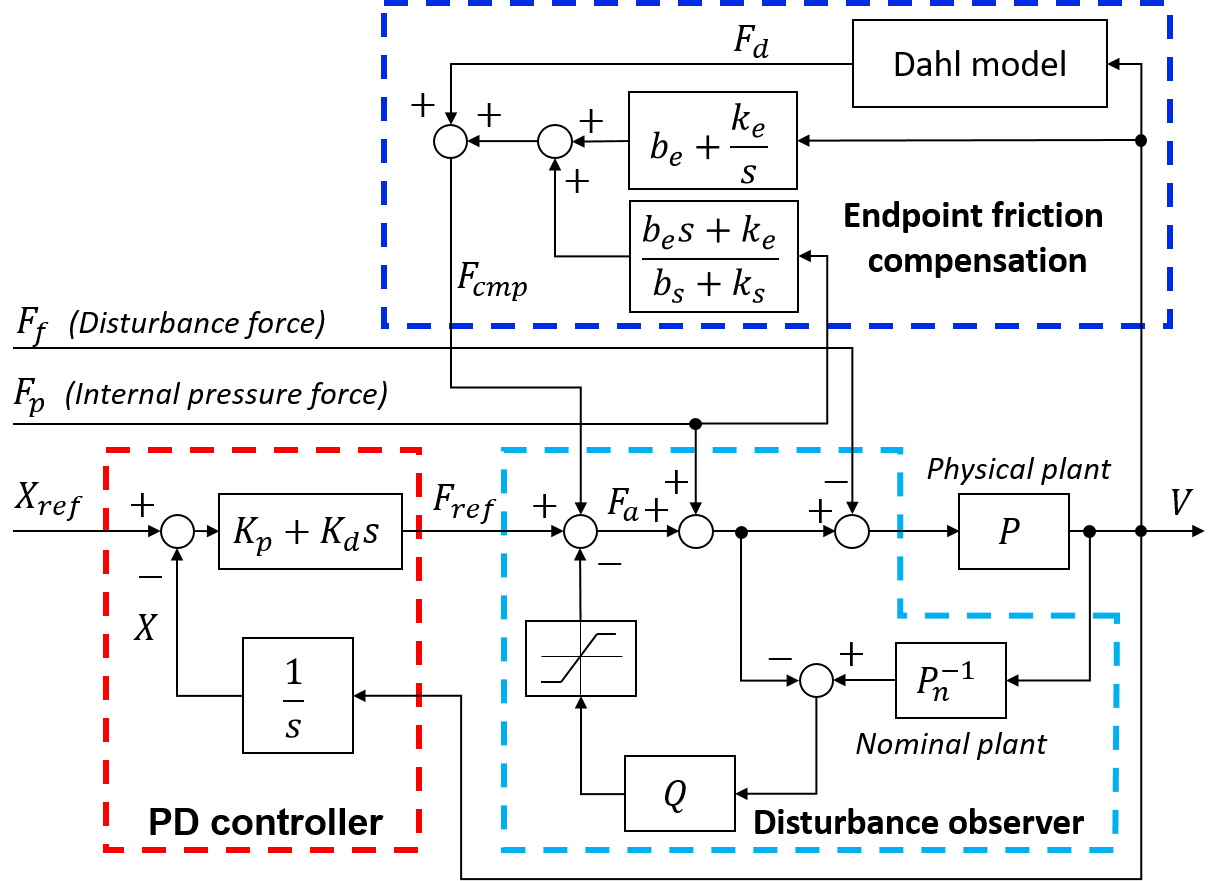}}}
    \caption{Disturbance observer for 1-DOF actuator motor plant with PD controller and model-based feed-forward friction compensation. The disturbance is estimated by comparing force input and velocity output times a nominal plant. The model-based friction compensation contains an approximated linear part and a nonlinear Dahl model part.}
    \label{fig:control framework in force feedback and DOB}
\end{figure}

For a fluid actuated manipulator, we use the internal transmission fluid pressure to measure internal force; since internal force feedback of a 2-DOF system is equivalent to external force feedback of 1-DOF motor system, we start by applying the DOB to the motor plant (Fig.~\ref{fig:control framework in force feedback and DOB}). The sum of forces on the motor is
\begin{equation}
    \label{eqn:DOB}
    F_a(s)+F_p(s)-F_f(s)=P^{-1}(s)V(s),
\end{equation}
where $P$ is the motor plant. From the controller block diagram (without feed-forward compensation term here),
\begin{equation}
    \label{eqn:actuator force of DOB}
    \begin{split}
        F_a(s)= & \frac{1}{1-Q(s)}F_{ref}(s)+\frac{Q(s)}{1-Q(s)}F_p(s) \\ 
        & -\frac{Q(s)}{1-Q(s)}P_n^{-1}(s)V(s),
    \end{split}
\end{equation}
where $F_{ref}$ is the reference force (e.g. set by an outer-loop position controller), $P_n$ is the nominal plant, and $Q$ is a low-pass filter. 
The order of $Q$ depends on the order of motor plant. Since the inverse of nominal plant is $P_n^{-1}=m_ns+b_n+k_n/s$, $Q$ must be at least first order to make $Q P_n^{-1}$ a proper transfer function~\cite{shim_yet_2016}.  Thus, we choose $Q(s)=\lambda/(s+\lambda)$,
where $\lambda$ is the cut-off frequency of low-pass filter. Since human input frequency is from $4Hz$ to $8Hz$, we can set cutoff frequency to $20Hz$ for the low-pass filter to limit the uncertainty or noise of system and maintain low frequency interaction at the same time. This is inspired from Fig.~\ref{fig:system ID}-A, where the confidence region starts to expand at around $150rad/s$, which is about $23.8Hz$. In low frequency limit ($\omega<\lambda$), $Q\approx 1$. Combining Eqs.~\ref{eqn:DOB}--\ref{eqn:actuator force of DOB} and 
assuming for low frequency range, the expression for velocity can be written as
\begin{equation}
    \label{eqn:velocity output of DOB at low frequency}
    V(s)=P_n(s)\left(F_{ref}(s)+F_p(s)\right),
\end{equation}
and so, the closed-loop system dynamics approach the nominal plant.  By selecting a frictionless nominal plant $P_n=1/ms$, rather than the best linear approximation of the actual plant, the DOB can theoretically compensate for all internal friction without modifying the system inertia.

To render minimum endpoint impedance, our reference/target force is $F_{ref}=0$; 
then Eq.~\ref{eqn:actuator force of DOB} simplifies to 
\begin{equation}
    \label{eqn:simplified integral form of dob}
    F_a(s) = \frac{\lambda}{s}(F_p(s) - P_n^{-1}(s)V(s)).
\end{equation}
We can think of $P_n^{-1}$ as a target impedance, which in the classic force-feedback example, would be set to zero, and the control signal is thus the integral of the difference between the measured force and the force predicted from a target (zero) impedance, $F_a(t) = \lambda\int_0^t F_p(\tau)\textrm{d}\tau$; integral force feedback is very commonly used in force-feedback systems~\cite{raibert1981hybrid}.  The case where a target impedance is $P_n^{-1}=ms$ in the force-feedback framework is equivalent to proportional acceleration feedback from Eq.~\ref{eqn:simplified integral form of dob}.

\subsection{Passivity Requirement}
The passivity criterion is a useful tool to check the overall contact or coupled stability with any passive system. We can use it to determine the possible range of parameters of nominal plant $P_n$ over which the closed-loop system will remain stable, when coupled to any passive environment. A linear time-invariant 1-port is passive if and only if (i) its port admittance $Y(s)$ (or impedance $Z(s)$) has no poles in the right half plane, (ii) imaginary poles of $Y(s)$ are simple with positive real residues, and (iii) $Re{\{Y(j\omega)\}}\geq 0$~\cite{colgate_control_1988}.  To check passivity criterion of the disturbance observer, we
use an equivalent parallel stiffness and damping to represent friction for 1-DOF motor plant.   The admittance of system considering DOB is shown as:
\begin{equation}
    \begin{split}
        Y(s) & =\frac{V(s)}{F_p(s)}=\frac{1}{\left(1-Q(s)\right)P^{-1}(s)+Q(s)P_n^{-1}(s)} \\
        & =\frac{s\left(s+\lambda\right)}{ms^3+\left(\lambda m_n+b\right)s^2+\left(k+\lambda b_n\right)s+\lambda k_n},
    \end{split}
    \label{eqn:admittance}
\end{equation}
where $m_n$, $b_n$ and $k_n$ are the nominal mass or inertia, nominal damping and nominal stiffness of 1-DOF motor plant. If there are no poles on right half plane (RHP), $m_n\geq -b/\lambda$, $b_n\geq -k/\lambda$ and $0\leq k_n\leq (\lambda m_n+b)(\lambda b_n +k)/(\lambda m)$ are needed by Routh-Hurwitz criterion. For the (ii) criterion, we need only to consider the range of $m_n$, $b_n$ and $k_n$ based on the (i) criterion. If there is only one pole on imaginary axis, which requires $b_n=-k/\lambda$ and $k_n=0$, the residue is $\lambda/(\lambda m_n+b)$, which is positive from the range of $m_n$ based on (i) criterion. If two conjugate imaginary poles exist, which requires $m_n=-b/\lambda$, $k_n=0$ and $b_n\neq -k/\lambda$, the real part of residue equals $0.5$, which is also positive. If two poles are located at origin, which requires $m_n= -b/\lambda$, $b_n= -k/\lambda$ and $k_n=0$, this is the only situation that contradicts the (ii) criterion because of repeated non-simple poles on the imaginary axis. Then, based on (i) and (ii) criterion, we have $m_n> -b/\lambda$, $b_n\geq -k/\lambda$ and $0\leq k_n\leq (\lambda m_n+b)(\lambda b_n +k)/(\lambda m)$. The real part of $Y(j\omega)$ is written as:
\begin{equation}
    \begin{split}
        & Re\left(Y(j\omega)\right)= \\
        & \frac{\lambda\omega^2\left(k+\lambda b_n-k_n\right)+\omega^4\left(b+\lambda m_n-\lambda m\right)}{\left(\lambda k_n-\left(b+\lambda m_n\right)\omega^2\right)^2+\left(\left(k+\lambda b_n\right)\omega-m\omega^3\right)^2}.
    \end{split}
\end{equation}
Based on the (iii) criterion, we can get $k_n\leq k+\lambda b_n$ and $m_n\geq m-b/\lambda$. Then the overall passivity criterion requires 
\begin{equation}
m_n\geq m-b/\lambda \quad\textrm{and}\quad 0\leq k_n\leq k+\lambda b_n.
\end{equation}
Note that the series dynamics and endpoint inertia, damping, and friction may all be considered part of the passive environment coupled to the motor plant, so the passivity result holds in the series-elastic 2-DOF case.

The nominal plant inertia $m_n$ is limited to $m_n \geq m-b/\lambda$ to maintain passivity, while the nominal damping $b_n$ and nominal stiffness $k_n$ can be reduced to zero.  Moreover, $b_n$ can be even set to a small negative number larger than $-k/\lambda$.  In the present case for our gripper, $\lambda$ is about 10 times larger than the ratio of $b/m$, which means the lower bound of $m_n$ is very closed to $m$. We can set $m_n=m$ for a conservative consideration.  It's possible to set a low value of $\lambda$ to allow a reduction of the driving point mass (small $m_n$), but this will undesirably bandlimit the friction compensation effect.

Notice that a traditional integral force feedback controller will be non-passive, unless very low integral force feedback gain is used ($\lambda$ small) or the system has high physical damping ($b$ large), as such a controller chooses $m_n=0$ implicitly.

The passivity criterion is a conservative requirement for contact stability or coupled stability, but violating these limits must be done carefully with experimental testing over all potential contact conditions.



\section{FEED-FORWARD ENDPOINT FRICTION COMPENSATION}
\begin{figure}[thpb]
    \centering
    \framebox{\parbox{3in}{\includegraphics[width=3in]{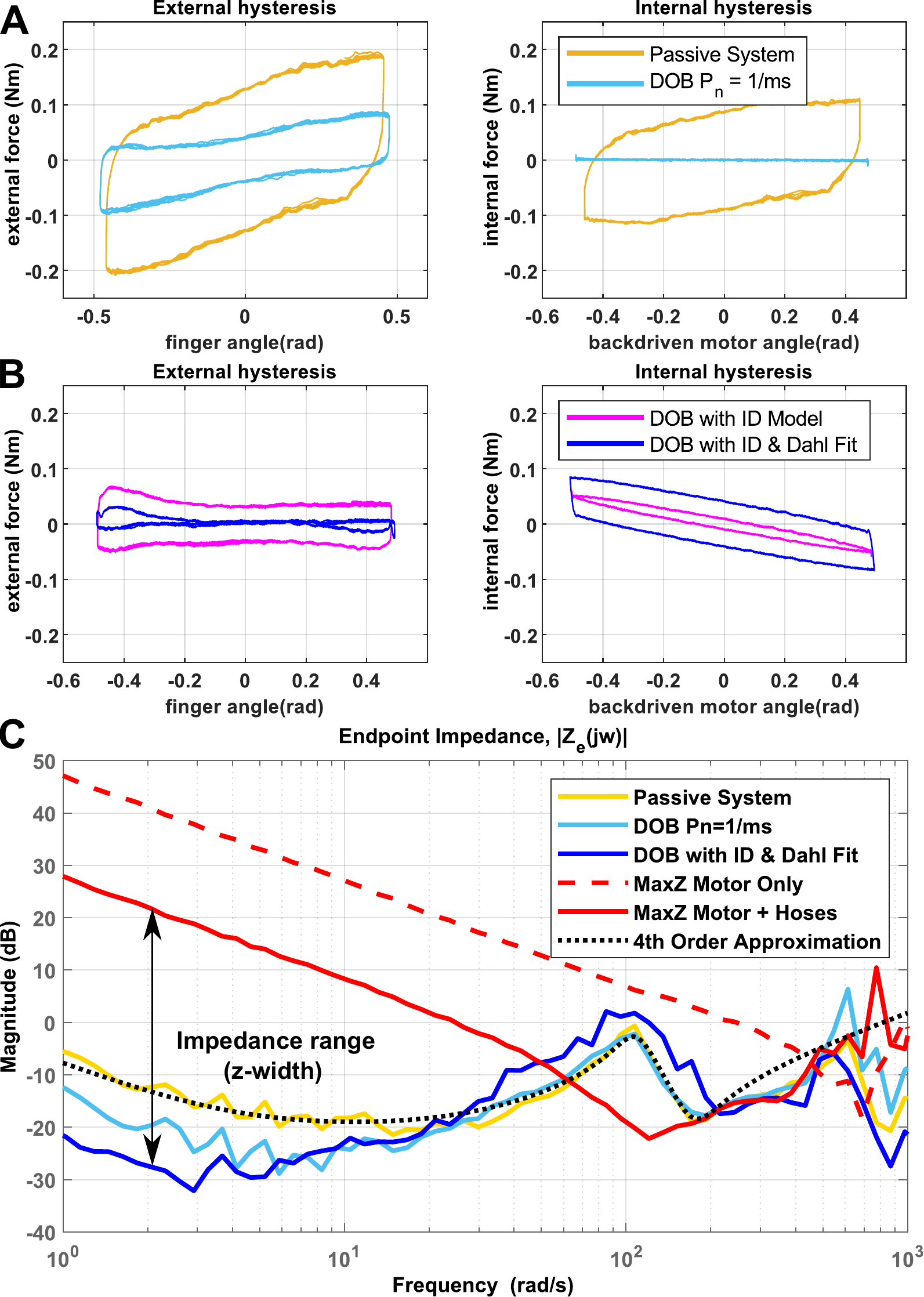}}}
    \caption{Experimentally measured gripper performance with force feedback. \textbf{(A--B)} compare the external pressure force and internal pressure force (torque equivalent) work loops for a quasi-statically backdriven finger ($\omega=1$~rad/s, amplitude =~0.5~rad).  {\bf(A)} Hysteresis of passive system versus DOB controller.  The DOB effectively removes all \emph{internal} friction, damping, and hysteresis (right), but not at the endpoint (left). {\bf(B)} DOB controller with feedforward term for only identified endpoint linear damping and stiffness (magenta); with nonlinear Dahl hysteresis compensation included. {\bf(C)} Identified endpoint impedance $Z(s)=F_e(s)/V_e(s)$ with different controllers.  Z-width comparison is made against the case (red) where the backdriven motor is commanded to hold with maximum stable PD gains.}
    \label{fig:hysteresis}
\end{figure}

Without external force sensing, model-based friction compensation~\cite{yao_adaptive_2015} can be used at the endpoint. In this section, we separate the friction compensation into the linear part (i.e. any parallel endpoint stiffness and damping, $k_e$ and $b_e$, found through previous system ID) and nonlinear part, found by observing hysteresis in a quasi-static external force work loop.  The contribution from motor friction is almost entirely compensated by the DOB (see cyan loops in Fig.~\ref{fig:hysteresis}-A).  We use an identified linear model of the transmission and endpoint friction to use motor state and internal pressure measurements to observe the endpoint state (velocity, $V_e$), which combined with the identified endpoint friction model offers an estimate of the endpoint friction $F_g$.
This ``friction feed-forward'' approach is shown schematically in Fig.~\ref{fig:control framework in force feedback and DOB}.


The Laplace-transformed endpoint compensation equation can be written as $F_{cmp}=\left(b_e+k_e/s\right)\hat{V_e}$, where $\Hat{V_e}$ is the estimated endpoint velocity based on identified hydraulic line model using internal pressure force $F_p$ and motor velocity $V$. It can be shown as $\hat{V_e}=V+F_p/\left(b_s+k_s/s\right)$. Substituting this $\hat{V_e}$ into the endpoint compensation equation,
\begin{equation}
    \label{eqn:finger damping and stiffness compensation using hydraulic plant}
    \begin{split}
        F_{cmp}(s) & =\left(b_e+\frac{k_e}{s}\right)\left(V(s)+\frac{1}{b_s+\frac{k_s}{s}}F_p(s)\right) \\
        & =\left(b_e+\frac{k_e}{s}\right)V(s)+\frac{b_es+k_e}{b_ss+k_s}F_p(s).
    \end{split}
\end{equation}

\begin{figure}[thpb]
    \centering
    \framebox{\parbox{3in}{\includegraphics[width=3in]{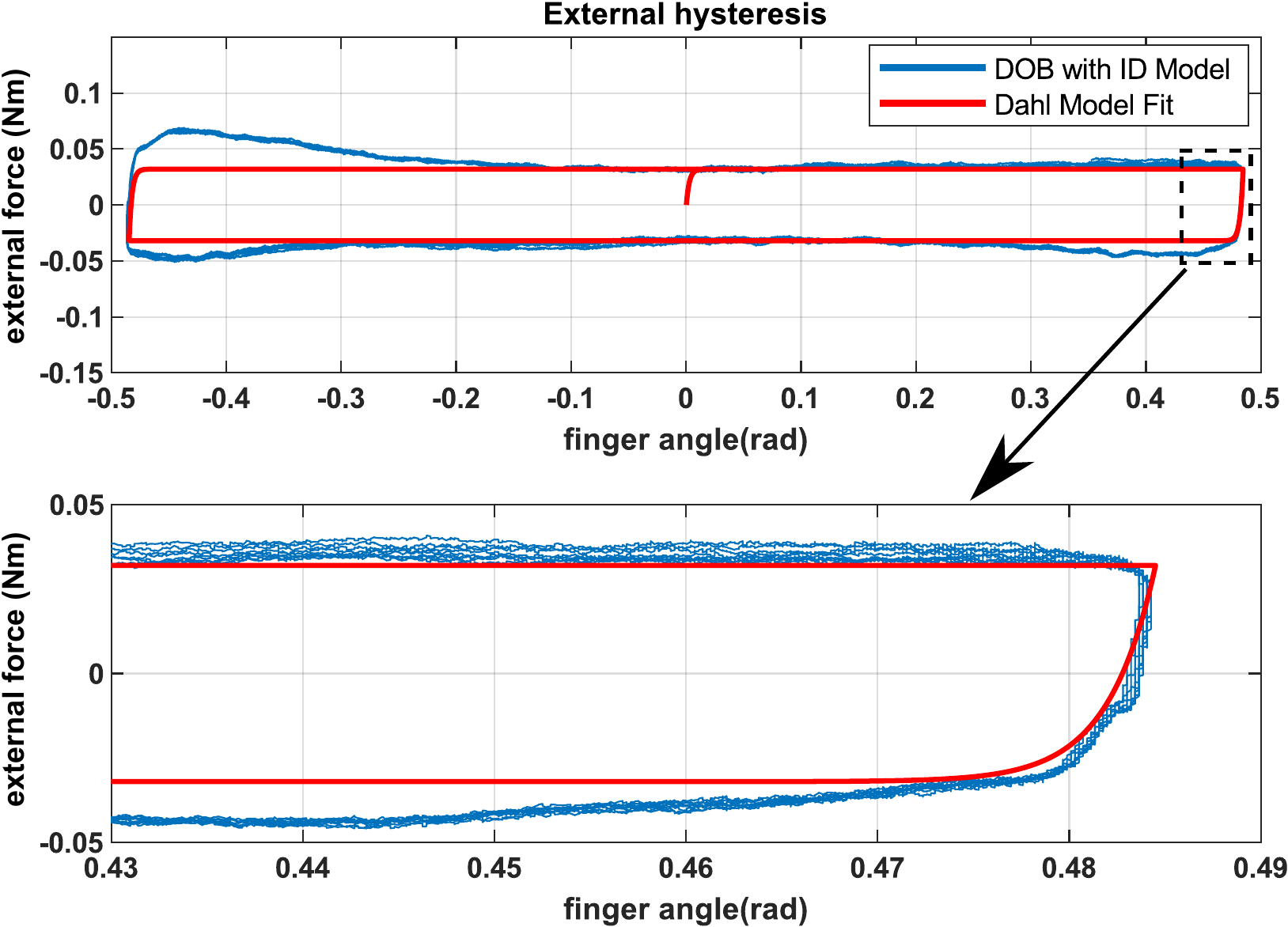}}}
    \caption{Dahl model fit for external hysteresis loop using simplified Dahl model $F_c=0.032Nm$ and $\sigma=12.8Nm/rad$ with close-up view of hysteresis loop in vicinity of turnaround point.}
    \label{fig:Dahl}
\end{figure}

After applying this compensation, the internal pressure force (cyan curve), rather than following/targeting zero (Fig.~\ref{fig:hysteresis}-A), now follows the \emph{negative of the identified linear endpoint stiffness and damping}, aiming to cancel them out. Fig.~\ref{fig:hysteresis}-B shows the result, where only the identified linear model is used (magenta curve); the remaining friction appears to be a pure Bouc-Wen type hysteresis~\cite{bouc-wen}, which is similar to the Dahl friction model~\cite{dahl}. This friction/hysteresis force, $F_d$, evolves according to
\begin{equation}
    \label{eqn:gernal Dahl model}
    \frac{dF_d}{dx}=\sigma\left|1-\frac{F_d}{F_c}\sgn{\left(\dot{x}\right)}\right|^n\sgn{\left(1-\frac{F_d}{F_c}\sgn{\left(\dot{x}\right)}\right)},
\end{equation}
where $\sigma$ is a stiffness parameter at equilibrium ($F_d=0$~Nm); $F_c$ is the hysteresis amplitude; $n$ is a
material dependent parameter which is $0\leq n\leq 1$ for brittle materials and $n\geq 1$ for ductile materials~\cite{van_geffen_study_2009}. For simplicity, $n=1$, the usual choice in friction modeling, resulting in the simplified form, $\textrm{d}F_d/\textrm{d}t=\sigma\dot{x}\left(1-(F_d/F_c)\sgn{\left(\dot{x}\right)}\right)$.  This two-parameter model is fit to the external hysteresis loop~(Fig.~\ref{fig:Dahl}). $F_c=0.032$~Nm and $\sigma=12.8$~Nm/rad are estimated for this model.

We include this nonlinear correction by adding $F_d$ to the right hand side of Eq.~\ref{eqn:finger damping and stiffness compensation using hydraulic plant}. The external force within the range $\left(-0.2,0.2\right)$~rad is reduced to almost zero (Fig.~\ref{fig:hysteresis}-B and Fig.~\ref{fig:Dahl}). Outside this range, there is still a small amount of uncompensated backdrive force due to a nonlinear four-bar mechanism gear ratio in the finger joint, which is unaccounted for, and a slight stroke position dependence on the actuator diaphragm resistance force~(Fig.~\ref{fig:Dahl}).  A more sophisticated visco-elastic model of the diaphragm could be used to better fit the observed behavior~\cite{slightam2012novel}.

Fig.~\ref{fig:hysteresis}-C shows the identified endpoint impedance, with and without force feedback. The DOB controller ($P_n^{-1}=ms,\;\lambda=20\textnormal{~rad/s}$) reduces endpoint impedance $|Z(j\omega))|$ by 7--10dB up to 10rad/s compared to the passive case. The DOB controller with linear + Dahl friction feed-forward reduces endpoint impedance up to 17dB, while \emph{increasing} impedance up to 6dB over the passive case at higher frequencies.  This makes sense, as the hysteresis model was calibrated only under quasi-static conditions.  A dynamic hysteresis model for fluid actuators (e.g.~\cite{yao_adaptive_2015}) could be used to better compensate for hysteresis in mid-range frequencies.

Z-width is calculated by comparing force-feedback targeting zero impedance (finger most backdrivable) to a stiff PD controller attempting to clamp the backdriven motor (maximum finger stiffness, red line). As shown in Fig.~\ref{fig:hysteresis}-C, the Z-width range that can be rendered is $\sim 50$~dB up to $3$~rad/s and $>30$~dB up to $10$~rad/s.  The dashed red line (``MaxZ Motor Only'') is the maximum impedance measured at the motor output port ($F_p(s)/V(s)$), representing the maximum endpoint impedance \emph{if} rigid hoses were used instead of rubber hoses here, which have modest volumetric compliance.  In the rigid-hose case, the theoretical maximum Z-width is 70~dB at DC.

\paragraph{Accompanying Video}
The supplemental video shows (i) nominal operation of the gripper with mid-range PD gains; (ii) passive backdrivability with motors off (impedance shown as yellow curve in Fig.~\ref{fig:hysteresis}-C); (iii) backdriving with force feedback (active zero impedance mode, impedance shown as blue curve in Fig.~\ref{fig:hysteresis}-C), showing that a seagull feather and $0.5$~mm mechanical pencil lead (\emph{pushing sideways}) can backdrive the gripper finger (the backdrive force at the fingertip is within $\pm5$~grams over the middle $10$~cm of the fingertips' linear range); (iv) the gripper crushes an empty aluminium beverage can under max PD gains (impedance shown as red curve in Fig.~\ref{fig:hysteresis}-C).

\paragraph{Safety}  If a \emph{passively safe} robot is desired, then it is important to size the motors such that the maximum speed achieved under peak torque output is within safe limits, given the moving mass and surface hardness of the manipulator.  Over-sizing motors increases the maximum forces that may be applied, but speeds must be monitored and electronically limited to preserve \emph{active safety}\footnote{For example, ISO~10218 and TS/15066 safety standards for industrial collaborative robots.}.  In both approaches, fluid-driven soft actuators can greatly reduce the endpoint mass, improving safety, and accurate friction compensation allow for more precise operation when motor gains are low, facilitating interaction with delicate environments.

\section{CONCLUSIONS}

Developing lightweight robotic manipulators that can offer both a high maximum stiffness and a low minimum endpoint impedance (low mass, low friction) is challenging; the design of low-friction soft actuators and the use of closed-loop force feedback are both effective tools to improve the endpoint impedance range.  Increasing endpoint backdrivability without external force sensing is especially challenging.  In the case of fluid-driven soft actuators that leverage internal pressure force feedback, we find the following:
\begin{itemize}
    \item Discontinuous (Coulomb) friction and deadband (backlash) at the endpoint, being entirely unobservable from internal force, must be minimized to the greatest extent possible through actuator topology and material design; hysteresis and viscous friction should be minimized, but they can also be reduced by active compensation, \emph{even without endpoint state or external force feedback}.
    \item Endpoint friction (e.g. material hysteresis) compensation is feasible if the friction is repeatable and can be captured accurately with a suitable model.  Adaptive control, not employed in this work, could be used if system parameters drift with time.
    \item Estimation of the endpoint state in fluid-actuated systems without explicit endpoint sensing is degraded by non/poorly-backdrivable actuators since accurate endpoint state estimation depends on internal pressure and driving-point state feedback.
    \item Admittance control (force feedback) is helpful even under the nearly-ideal situation of a direct-drive brushless motor actuator, but it is an open question as to the continued effectiveness of this approach as increasingly higher impedance torque sources (e.g. smaller motors with a high gear ratio) are employed.
\end{itemize}

There are many future steps to improve and extend this work.  Consideration of the noise and resolution limits of the actuator position and force sensors can better inform the achievable closed-loop impedance range and stability limits.  This is critical for applying the techniques in this paper to pneumatically actuated systems, where the low series stiffness makes it much harder to reconstruct endpoint state from internal pressure signals.

Another exciting possible extension is to add an online observer to estimate/identify the environment/interaction impedance continuously.  This ability would be useful for measuring the stiffness and damping properties of grasped objects, and useful for adaptively tuning manipulator impedance on-the-fly to the optimal value, given the continuously varying environmental impedance.









\section*{ACKNOWLEDGMENT}

This work is funded in part under NSF NRI 1830425.



\bibliographystyle{IEEEtran}
\bibliography{bibtex/bib/IEEEabrv.bib,bibtex/bib/IEEEexample.bib,bibtex/bib/IEEEpaper.bib}{}



\end{document}